\documentclass[conference]{IEEEtran}
\IEEEoverridecommandlockouts
\usepackage{cite}
\usepackage{amsmath,amssymb,amsfonts}
\usepackage{graphicx}
\usepackage{booktabs}
\usepackage{url}
\usepackage{balance}
\usepackage{tikz}
\usepackage{pgfplots}
\usepackage{hyperref}
\pgfplotsset{compat=1.18}
\usetikzlibrary{arrows.meta,positioning}
\usepackage{placeins}
\usepackage{algorithm}
\usepackage{algpseudocode}

\begin{document}

%\title{Detecting Temporal Inconsistencies in Political News through Temporal Knowledge Graphs and an Explainable Coherence Score}
\title{An Explainable Coherence Score for Detecting Temporal Inconsistencies in Political News}
% Double-blind submission: author information intentionally omitted.

\author{Marius Nicusor Pantea and Adrian Groza \\ Artificial Intelligence Research Institute AIRi@UTCN \\ Technical University of Cluj-Napoca, Cluj-Napoca, Romania \\
  marius.pantea@campus.utcluj.ro, adrian.groza@cs.utcluj.ro}

\maketitle

\begin{abstract}
Temporal inconsistencies, such as mandates attributed outside their real interval, events presented as past before they occurred, or inverted causal sequences, are a form of political disinformation that evades style-based fake news detectors: a well-written article with a single wrong date carries no lexical signal of falsehood. This paper introduces the Temporal Coherence Score (TCS), a continuous, intrinsically interpretable metric that quantifies the temporal coherence of a news article, computed by a four-stage pipeline: extraction of temporal facts, construction of a temporal knowledge graph, hierarchical verification against internal consistency rules and external reference sources, and score aggregation with automatically generated explanations. Verification combines eight internal checkers derived from Allen's interval algebra with a five-level external hierarchy ranging from a locally stored reference knowledge base of 1{,}256 curated political facts to live Wikidata SPARQL queries. On a benchmark of 100 political news articles with injected temporal errors, the system reaches a precision of 0.909 at the selected operating threshold, with a single residual false positive, a profile deliberately tuned for human-in-the-loop fact-checking assistance, where false alarms are costlier than missed detections. Unlike lexical baselines that output only a binary label, every flagged article is accompanied by the inconsistency type, the entities involved, and the reference source that contradicts the claim.
\end{abstract}


\section{Introduction}
Automated fake news detection has largely concentrated on linguistic style, source credibility, and propagation patterns~\cite{zhou2020survey}. These families of methods share a blind spot: an article written in neutral language, published by a credible outlet, and quoting real sources may still contain a factual error that no stylistic feature can reveal \cite{11479686}. Temporal errors are a prominent instance. A text stating that a politician signed a law in 2004 when his mandate ended in 2001, or presenting a treaty as preceding the events it responded to, is grammatically and stylistically indistinguishable from an accurate report. Readers tend to accept presented chronologies without independent verification, and manual fact-checking organizations can only cover a small fraction of the daily output of political news.

Existing research on temporal fact verification treats temporal errors as a special case of claim verification, operating on isolated statements~\cite{qudus2023temporalfc,tsverver2024,chronofact2025}. Two gaps remain. First, no dedicated metric quantifies the temporal coherence of an entire article, aggregating the verification of multiple interdependent claims into a single interpretable value. Second, explainability is rarely a first-class requirement: a binary verdict without a verifiable justification cannot be integrated into an editorial workflow.

This paper addresses both gaps with the following contributions:
\begin{itemize}
\item the \emph{Temporal Coherence Score} (TCS), a continuous metric in $[0,1]$ that aggregates severity-weighted temporal inconsistencies, internal graph coherence, and external verification coverage, and is interpretable by construction: each factor of the formula corresponds to a well-defined semantic concept;
\item an end-to-end system that computes TCS through a four-component pipeline (extraction, temporal knowledge graph construction, hierarchical verification, scoring and explanation), running fully on-premise for inference and generating natural-language explanations with a compact local language model;
\item an evaluation on a purpose-built benchmark of 100 political news articles with five categories of injected temporal errors, including a threshold analysis, a comparison against lexical baselines, and an explicit delimitation of the domain of applicability on two external datasets.
\end{itemize}

\section{Related Work}
Zhou and Zafarani~\cite{zhou2020survey} organize fake news detection into knowledge-based, style-based, propagation-based, and source-based methods. Style-based classifiers perform strongly on datasets whose false class carries distinctive lexical patterns, but by design cannot detect factual errors in stylistically neutral text; knowledge-based verification against structured sources addresses precisely this vulnerability. Standard resources such as LIAR~\cite{wang2017liar} and FEVER~\cite{thorne2018fever} operate at the level of short isolated claims and provide no temporal context at article level; FakeNewsNet~\cite{shu2020fakenewsnet} adds social and spatio-temporal context but no temporal verification mechanism.

Temporal fact verification has recently emerged as a distinct direction. TemporalFC~\cite{qudus2023temporalfc} predicts the validity interval of a claim using temporal knowledge graph embeddings, but verifies claims in isolation and produces neither an article-level score nor explanations. TSVer~\cite{tsverver2024} benchmarks verification against time-series evidence, exposing the weakness of current systems in numerical-temporal reasoning. ChronoFact~\cite{chronofact2025} verifies claims involving multiple, overlapping, or recurring events by constructing event timelines, again at claim level. Allein et al.\ study the temporal ordering of evidence~\cite{allein2021timeaware} and implicit temporal reasoning~\cite{allein2023implicit}, showing that ignoring implicit temporal constraints causes systematic classification errors for entities with well-defined mandates. Soulard et al.~\cite{soulard2025temporalconstraints} validate knowledge-graph facts against automatically discovered temporal constraints grounded in Allen's interval algebra~\cite{allen1983maintaining}, with each decision justified by the violated constraint; the present work adopts the complementary strategy of a fixed set of checkers derived from the inconsistency types relevant to political text.

On the explainability side, Ngueajio et al.~\cite{ngueajio2025decoding} survey XAI techniques for fake news detection and identify human-in-the-loop integration as underexplored. Coroama and Groza~\cite{coroama2022evaluation} argue for evaluation of explainability at the level of individual decisions, a perspective that directly motivated the design of TCS as a decomposable score. MindBugs~\cite{cheres2025mindbugs} is an explainable disinformation detection system with a human-in-the-loop mechanism operating on narrative structures; it works at article level and produces explanations, but does not address the temporal dimension. Recent agentic and causal approaches~\cite{cui2026t2agent,wang2025explainablefakenews,local2025factchecking} generate structured justifications but likewise lack an article-level temporal coherence metric.

A related but architecturally distinct alternative is prompting a general-purpose large language model directly to judge the temporal plausibility of a claim, relying on its parametric knowledge rather than an explicit, auditable reference source. Such an approach offers no verifiable provenance for a verdict and is prone to hallucinated dates and mandates, particularly for less prominent political figures; the present system instead grounds every external check in a queryable source (a curated reference base or live Wikidata), so that a verdict can always be traced to a specific, inspectable fact.

To the best of our knowledge, no existing system combines article-level input, verification over a temporal knowledge graph, explanation generation, and an aggregated interpretable score; this combination is the contribution of the present work.

\section{Method}

\subsection{Problem Formalization}
Given an article $\mathcal{A}$, the system extracts a set of temporal facts $\mathcal{F}=\{(e_i,r_i,v_i,t_i)\}_{i=1}^{n}$, where $e$ is a political entity, $r$ a relation type, $v$ the associated value, and $t$ a temporal anchor expressed as a point $t_p$ or an interval $[t_s,t_e]$. Five relation types are used, selected by frequency in political news: \texttt{HOLDS\_POSITION}, \texttt{OCCURRED\_ON}, \texttt{MEMBER\_OF}, \texttt{FOLLOWED}, and \texttt{GENERIC}. Facts are verified against a reference knowledge base $\mathcal{K}$ and against each other, and the outcome is aggregated into a score $\mathrm{TCS}\in[0,1]$; an article is classified as temporally suspect if $\mathrm{TCS}<\theta$, with $\theta$ determined empirically.

The system targets five categories of temporal inconsistency: mandate errors (a position attributed outside its real interval, e.g., a former minister quoted as still holding office after resignation), anachronisms (an event or policy referenced as already in effect before its actual date), causal violations (an effect placed before its cause, e.g., a resignation dated earlier than the scandal that prompted it), temporal cycles in precedence chains (entity A reported as preceding B while B is separately reported as preceding A), and impossible simultaneous roles (the same individual holding two mutually exclusive positions at once).

\subsection{Pipeline Architecture}
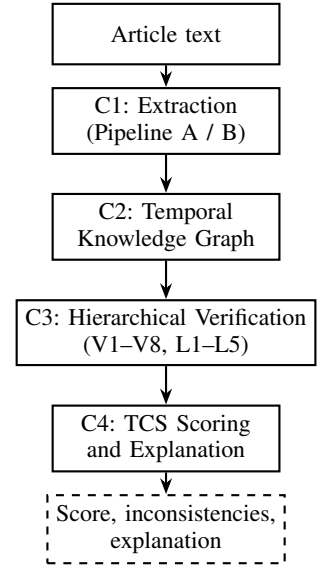
\begin{figure}[t]
\centering
\begin{tikzpicture}[
  box/.style={rectangle, draw, thick, minimum width=3.0cm, minimum height=0.8cm, align=center, font=\small},
  outbox/.style={rectangle, draw, thick, dashed, minimum width=3.0cm, minimum height=0.9cm, align=center, font=\small},
  arr/.style={-{Stealth[length=6pt]}, thick}
]
\node[box] (in) at (0,0) {Article text};
\node[box] (c1) at (0,-1.15) {C1: Extraction\\(Pipeline A / B)};
\node[box] (c2) at (0,-2.55) {C2: Temporal\\Knowledge Graph};
\node[box] (c3) at (0,-3.95) {C3: Hierarchical Verification\\(V1--V8, L1--L5)};
\node[box] (c4) at (0,-5.35) {C4: TCS Scoring\\and Explanation};
\node[outbox] (out) at (0,-6.55) {Score, inconsistencies,\\explanation};
\draw[arr] (in) -- (c1);
\draw[arr] (c1) -- (c2);
\draw[arr] (c2) -- (c3);
\draw[arr] (c3) -- (c4);
\draw[arr] (c4) -- (out);
\end{tikzpicture}
\caption{Four-stage TCS pipeline: extraction, knowledge graph construction, hierarchical verification, and scoring with explanation.}
\label{fig:architecture}
\end{figure}

The pipeline consists of four sequential components with strict separation of responsibilities, exposed through a REST API and a web interface. Figure~\ref{fig:architecture} summarizes the flow from raw article text to the final score and explanation.

\textbf{C1 --- Extraction.} Two alternative extraction strategies are provided. Pipeline~A is deterministic: a transformer-based NLP model~\cite{spacy2020} performs named entity recognition and dependency parsing; subject--predicate--object triples with temporal anchoring are derived from the dependency tree through a cascade of four strategies with decreasing confidence (dependency-based extraction, reflexive event facts, nominal association, and an entity--date fallback), and a verb ontology maps root lemmas to relation types. Pipeline~B uses a compact generative language model (1.7B parameters~\cite{qwen3_2025}) with few-shot structured prompting and JSON-schema validation, recovering temporal claims expressed through implicit or periphrastic constructions that dependency analysis misses; sentences without a four-digit year are pre-filtered before any model call to bound latency. Temporal expressions from both pipelines are normalized to ISO~8601, with resolution of relative dates and approximate decade expressions.

\textbf{C2 --- Temporal knowledge graph.} Extracted facts are filtered by a minimum confidence threshold of $0.3$ and by the requirement of a valid temporal anchor, deduplicated on the exact tuple signature, and inserted into a directed multigraph following the TKG formalization $G=(E,R,T,F)$ with $F\subset E\times R\times E\times T$~\cite{cai2024tkgsurvey}. An optional graph database provides cross-article persistence and caching.

\textbf{C3 --- Hierarchical verification.} Verification combines two orthogonal mechanisms. Eight \emph{internal checkers} (V1--V8) operate on the graph alone, without network access, detecting: temporal cycles in precedence chains (V1), causal violations (V2), ordering errors and implausible durations (V3), the same position held simultaneously by different entities (V4), positions held before election or investiture (V5), future events presented as past (V6), incompatible simultaneous roles of one entity (V7), and actions performed before taking office (V8). Table~\ref{tab:examples} gives one illustrative example per checker. Algorithm~\ref{alg:c3} formalizes the order of application, from internal checkers to hierarchical external escalation.
%\ag{One example for each checker}
%\ag{Some algorithms for relevant checkers (with algorithmic package)}

\begin{table}[t]
\centering
\caption{One Illustrative Example per Internal Checker}
\label{tab:examples}
\footnotesize
\begin{tabular}{lp{6.3cm}}
\toprule
Code & Illustrative example \\
\midrule
V1 & Article states Blair \texttt{FOLLOWED} Brown, and separately Brown \texttt{FOLLOWED} Blair --- a cycle \\
V2 & A resignation dated before the scandal that caused it becomes public \\
V3 & A mandate spanning 95 years, exceeding plausible tenure \\
V4 & Two different entities both stated to hold ``President'' on the same date \\
V5 & A minister acting in office weeks before the confirming election \\
V6 & A 2030 policy referenced as already completed in a 2024 article \\
V7 & The same person presented as PM of two countries in the same month \\
V8 & An official signing a law before the recorded date of taking office \\
\bottomrule
\end{tabular}
\end{table}

\begin{algorithm}[t]
\caption{Hierarchical Verification}
\label{alg:c3}
\begin{algorithmic}[1]
\Require graph $G$, reference base $K_{L1}$, cross-article cache $K_{L2}$
\Ensure annotated inconsistency list $I$
\State $I \gets \Call{ApplyCheckers}{G}$ \Comment{V1--V8, internal, no network}
\ForAll{fact $f \in G$ unresolved by internal checkers}
  \If{$\Call{Verify}{f, K_{L1}}$} \Comment{L1 --- local reference KB}
    \State add result to $I$; \textbf{continue}
  \EndIf
  \If{$\Call{Verify}{f, K_{L2}}$} \Comment{L2 --- Neo4j cache}
    \State add result to $I$; \textbf{continue}
  \EndIf
  \If{$\Call{QueryWikidata}{f}$} \Comment{L3 --- SPARQL}
    \State add result to $I$; \textbf{continue}
  \EndIf
  \State $\Call{VerifyRSS}{f}$ \Comment{L5 --- recent facts}
  \State add result to $I$
\EndFor
\State $I \gets \Call{AggregateSeverities}{I}$ \Comment{LOW/MEDIUM/HIGH/CRITICAL}
\State \Return $I$
\end{algorithmic}
\end{algorithm}
The checkers instantiate relations of Allen's interval algebra~\cite{allen1983maintaining}, which defines thirteen mutually exclusive relations between two time intervals $I_1$ and $I_2$ --- \textit{before}, \textit{meets}, \textit{overlaps}, \textit{starts}, \textit{during}, \textit{finishes}, \textit{equals}, and the inverses of the first six. An \textit{overlaps} relation between two mandates of mutually exclusive offices held by the same entity signals V4/V7; a \textit{before} relation whose transitive closure over the \texttt{FOLLOWED} chain returns to its starting entity signals a temporal cycle (V1). The checkers further use explicit day-level tolerances (e.g., 30 days for inverted intervals, 180 days between election and investiture) to absorb the imprecision of dates extracted from text. \emph{External verification} escalates each externally verifiable fact through a hierarchy of sources: a locally stored reference knowledge base of 1{,}256 curated facts covering US, UK, and EU political mandates and canonical events (L1, zero latency); a persistent cross-article cache (L2, optional); live Wikidata SPARQL queries over the temporal qualifiers of position and membership properties~\cite{vrandecic2014wikidata} (L3). An identity guard compares the resolved Wikidata entity against the extracted name before accepting any result, preventing cross-entity confusion between similarly named individuals. For position-holding claims, verification is inverse rather than direct: all recorded positions and organizational memberships of the entity are retrieved from Wikidata, and the claimed interval is checked against every one of them, not only the specific position named in the article. If the claimed position is absent from the retrieved set entirely, the mismatch is flagged at reduced severity, since the gap may reflect incomplete Wikidata coverage rather than a genuine article error; if the position is present but under a conflicting interval, severity is high. This asymmetry reduces false positives caused by Wikidata's uneven coverage of less-documented offices. Verification continues, for unresolved entities, with a Wikipedia REST fallback (L4) that was tested and rejected after introducing false positives; and real-time RSS feeds for recent facts (L5, opt-in). A general tolerance of 200 days applies to external interval comparison, relaxed to year-level granularity when the article states only a year. Canonical-event matching applies five successive precision filters (minimum word count, action verb presence, fuzzy similarity, word overlap, and distinctive shared words) to avoid spurious matches. Table~\ref{tab:hierarchy} summarizes the domain, data source, and inconsistency-generating capability of each checker and level.

\begin{table}[t]
\centering
\caption{Internal Checkers and External Verification Levels}
\label{tab:hierarchy}
\footnotesize
\begin{tabular}{lp{4.0cm}p{1.3cm}}
\toprule
Code & Domain / data source & New incons.? \\
\midrule
V1--V8 & Internal TKG structure (cycles, ordering, contradictions, overlaps); in-memory graph & Yes \\
L1 & Reference mandates and canonical events; local file & Yes \\
L2 & Facts verified in prior articles; Neo4j (local) & Yes \\
L3 & Mandates and events from Wikidata; external SPARQL & Yes \\
L4 & Wikipedia verification (disabled in production); external REST & --- \\
L5 & Confirmation from recent news feeds; external RSS & No (confirm only) \\
\bottomrule
\end{tabular}
\end{table}

\subsection{Temporal Coherence Score}
Component C4 aggregates the verification outcome into
\begin{align}
\mathrm{TCS} = {}& (1-\text{penalty\_ratio}) \times \text{score\_coherence} \notag\\
& \times (0.85 + 0.15\times \text{coverage\_factor}),
\label{eq:tcs}
\end{align}
clipped to $[0,1]$. The penalty ratio is the severity-weighted sum of detected inconsistencies, normalized by the maximum possible penalty ($n_{\text{claims}}\times 1.5$), with weights $\{0.0, 0.3, 1.0, 1.5\}$ for severities LOW to CRITICAL. The coherence factor is the proportion of temporal relations unaffected by inconsistencies of MEDIUM severity or higher. The coverage factor is the fraction of facts verifiable against external sources; its multiplicative bonus is bounded to the interval $[0.85, 1.00]$, a deliberate design decision limiting the influence of external coverage to at most 15\% so that penalty and internal coherence dominate the score. Articles with no extractable temporal facts receive the neutral score 0.5, explicitly labeled as insufficient temporal data rather than a truth verdict. TCS is intrinsically interpretable: each factor corresponds to a named concept, and the classification can be justified by exhibiting the component values and the list of detected inconsistencies, without post-hoc approximation.

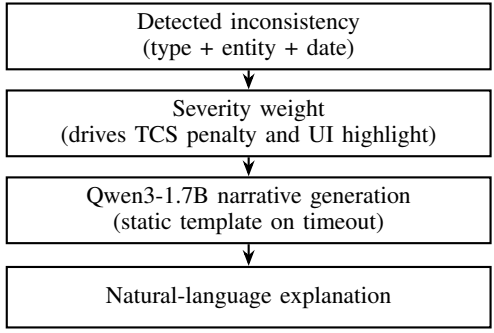
\begin{figure}[t]
\centering
\begin{tikzpicture}[
  box/.style={rectangle, draw, thick, minimum width=6.4cm, minimum height=0.8cm, align=center, font=\small},
  arr/.style={-{Stealth[length=6pt]}, thick}
]
\node[box] (det) at (0,0) {Detected inconsistency\\(type + entity + date)};
\node[box] (sev) at (0,-1.15) {Severity weight\\(drives TCS penalty and UI highlight)};
\node[box] (llm) at (0,-2.3) {Qwen3-1.7B narrative generation\\(static template on timeout)};
\node[box] (exp) at (0,-3.45) {Natural-language explanation};
\draw[arr] (det) -- (sev);
\draw[arr] (sev) -- (llm);
\draw[arr] (llm) -- (exp);
\end{tikzpicture}
\caption{Explanation-generation flow: detected severity drives both the TCS penalty and the UI highlight, while the inconsistency list feeds narrative generation, which falls back to a static template on timeout.}
\label{fig:xai}
\end{figure}

For each analysis, a compact local language model transforms the structured inconsistency list into a narrative explanation; if generation fails or times out, static per-type templates guarantee that every result carries a justification. Figure~\ref{fig:xai} summarizes this explanation-generation flow, the system's central explainability mechanism.

A human-in-the-loop mechanism activates once an analysis is persisted, exposing Confirm and Reject actions to the reviewer. The recorded verdict is stored as an attribute of the article node in the graph database; at the current stage it functions as an audit trail and as a substrate for future extensions of the verification logic, rather than being fed back into scoring, consistent with the design premise that the system assists, rather than replaces, human judgment.

All inference runs on commodity hardware without GPU acceleration. Pipeline A and the internal checkers (V1--V8) are lightweight and dominate neither latency nor cost; the local language model calls in Pipeline B and in explanation generation are the main contributors to per-article latency, since both rely on model inference rather than the rule-based path of Pipeline A. This places Pipeline A at an advantage for high-throughput batch processing, and Pipeline B at an advantage for coverage of implicitly phrased claims, motivating their availability as alternative, user-selectable strategies rather than a single fixed extractor.

\section{Evaluation}

\subsection{Benchmark and Protocol}
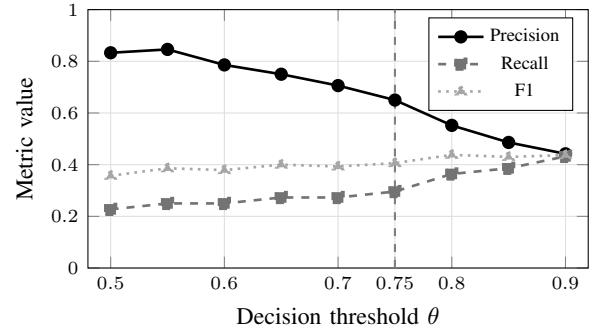
\begin{figure}[t]
\centering
\begin{tikzpicture}
\begin{axis}[
  width=8.2cm, height=5.0cm,
  xlabel={Decision threshold $\theta$},
  ylabel={Metric value},
  xmin=0.48, xmax=0.92, ymin=0, ymax=1,
  xtick={0.5,0.6,0.7,0.75,0.8,0.9},
  label style={font=\small},
  tick label style={font=\scriptsize},
  legend pos=north east, legend style={font=\scriptsize},
  grid=major, grid style={gray!25}
]
\addplot[black, mark=*, line width=1pt] coordinates {
  (0.50,0.833)(0.55,0.846)(0.60,0.786)(0.65,0.750)(0.70,0.706)(0.75,0.650)(0.80,0.552)(0.85,0.486)(0.90,0.442)};
\addplot[black!55, mark=square*, dashed, line width=1pt] coordinates {
  (0.50,0.227)(0.55,0.250)(0.60,0.250)(0.65,0.273)(0.70,0.273)(0.75,0.296)(0.80,0.364)(0.85,0.386)(0.90,0.432)};
\addplot[black!35, mark=triangle*, dotted, line width=1pt] coordinates {
  (0.50,0.357)(0.55,0.386)(0.60,0.379)(0.65,0.400)(0.70,0.393)(0.75,0.406)(0.80,0.438)(0.85,0.430)(0.90,0.437)};
\draw[dashed, gray, line width=0.8pt] (axis cs:0.75,0) -- (axis cs:0.75,1);
\legend{Precision, Recall, F1}
\end{axis}
\end{tikzpicture}
\caption{Precision, recall, and F1 across decision thresholds, measured on the intermediate development configuration used to select $\theta=0.75$ (Table~\ref{tab:evolution} traces the further corrections that raised final precision at this threshold to 0.909).}
\label{fig:sweep}
\end{figure}

Evaluation uses a purpose-built benchmark of 100 synthetic political news articles: 56 temporally coherent (TRUE) and 44 containing exactly one injected temporal error (FAKE), covering five inconsistency types (date mismatch, ordering error, future-as-past, entity inconsistency, implicit contradiction). TRUE articles describe chronologies validated against Wikidata; FAKE articles are derived from them by controlled error injection, each annotated with the injected type. An article is predicted FAKE if it yields at least one temporal fact and $\mathrm{TCS}<\theta$; articles without temporal facts are predicted TRUE. The operating threshold $\theta=0.75$ was selected through a sweep over $[0.50, 0.90]$ as the point that favors precision, since in the intended assistance context a false alarm against a legitimate source is costlier than a missed detection recoverable through manual review. Figure~\ref{fig:sweep} shows the sweep, measured on the development configuration used at the time of threshold selection; the subsequent precision-oriented corrections traced in Table~\ref{tab:evolution} raised precision at this same threshold to the final 0.909 reported in Table~\ref{tab:results}.

\subsection{Results}
Table~\ref{tab:results} reports the results at the operating threshold, and Table~\ref{tab:confusion} the confusion matrix. Precision is high: of the 11 articles flagged, 10 are truly fake, i.e., a single false alarm over the whole set. Precision at this operating point is computed from only 11 positive predictions; a 95\% Wilson score interval is $[0.62, 0.98]$, reflecting substantial uncertainty at this sample size and motivating evaluation on the larger 150-article benchmark constructed for future work. Accuracy (0.650) exceeds the majority-class baseline of 0.56. Recall is deliberately low (10 of 44 fakes detected), reflecting the precision-oriented operating point. The mean TCS of true articles is 0.880 versus 0.747 for fake articles, a separation of $+0.133$ whose partial overlap explains the missed detections.

\FloatBarrier

\begin{table}[t]
\centering
\caption{Results on the benchmark at $\theta=0.75$.}
\label{tab:results}
\begin{tabular}{lc}
\toprule
Metric & Value \\
\midrule
Precision & 0.909 \\
Recall & 0.227 \\
F1 score & 0.364 \\
Accuracy & 0.650 \\
\bottomrule
\end{tabular}
\end{table}

\begin{table}[t]
\centering
\caption{Confusion matrix at $\theta=0.75$.}
\label{tab:confusion}
\begin{tabular}{lcc}
\toprule
 & Predicted TRUE & Predicted FAKE \\
\midrule
Actual TRUE (56) & 55 & 1 \\
Actual FAKE (44) & 34 & 10 \\
\bottomrule
\end{tabular}
\end{table}

The reported configuration is the outcome of an iterative development process: starting from a baseline Wikidata verification with precision 0.714 and four false positives, targeted corrections (strict object matching, canonical-event verification, reduced date tolerance, role-category grouping, and the entity identity guard) raised precision to 0.909 and reduced false positives to one, each correction validated by re-running the benchmark. Table~\ref{tab:evolution} traces precision, recall, F1, and false-positive count across the five successive configurations.

\begin{table}[t]
\centering
\caption{Evolution of Performance During Development}
\label{tab:evolution}
\footnotesize
\begin{tabular}{lcccc}
\toprule
Configuration & P & R & F1 & FP \\
\midrule
Baseline verification & 0.714 & 0.227 & 0.345 & 4 \\
Extended tol.\ + cross-entity & 0.667 & 0.273 & 0.387 & 6 \\
Role-category grouping & 0.706 & 0.273 & 0.393 & 5 \\
FP correction ($\theta{=}0.70$) & 0.900 & 0.205 & 0.333 & 1 \\
Final ($\theta{=}0.75$) & 0.909 & 0.227 & 0.364 & 1 \\
\bottomrule
\end{tabular}
\end{table}

Two case studies illustrate the behavior. A factually accurate article on the Obama presidency yields six extracted facts, zero inconsistencies, and $\mathrm{TCS}=0.925$. A fake article stating that Bill Clinton signed a law in 2004, after his mandate ended in January 2001, yields four facts and six HIGH-severity date-mismatch inconsistencies against the Wikidata mandate interval, with $\mathrm{TCS}=0.0$.

\subsection{Comparison with Lexical Baselines}
Three lexical classifiers (logistic regression, linear SVM, random forest) over TF--IDF features (5{,}000 terms, uni- and bigrams) were evaluated by stratified 5-fold cross-validation on the same benchmark (Table~\ref{tab:baseline}). Random forest (F1 $=0.572$) and SVM (F1 $=0.535$) exceed the proposed system's F1 of 0.364, as expected when the false class carries lexical patterns. The advantage of the proposed system lies elsewhere: the baselines emit a bare binary label, whereas every TCS verdict identifies the inconsistency type, the entities involved, and the contradicting reference source, which is the functional requirement for editorial integration~\cite{coroama2022evaluation}.

\begin{table}%[t]
\centering
\caption{Lexical baselines on the benchmark (5-fold CV).}
\label{tab:baseline}
\begin{tabular}{lccc}
\toprule
Model & Precision & Recall & F1 \\
\midrule
Logistic Regression & 0.783 & 0.181 & 0.280 \\
Linear SVM & 0.777 & 0.436 & 0.535 \\
Random Forest & 0.701 & 0.500 & 0.572 \\
Proposed (TCS, $\theta{=}0.75$) & 0.909 & 0.227 & 0.364 \\
\bottomrule
\end{tabular}
\end{table}

\subsection{Domain Delimitation on External Datasets}\label{subsec:domain}
To characterize generalization, the system was run on 100 political articles each from ISOT~\cite{mogre2024isotliar} and RAGuard~\cite{raguard2026}. F1 collapses to 0.000 (ISOT) and 0.039 (RAGuard), with TCS class separations of $+0.065$ and $-0.052$ respectively. The causes differ and are informative: ISOT's fake class is characterized by emotional style rather than verifiable temporal errors, so the temporal signal that TCS measures is absent; RAGuard's short claims lack extended temporal context and frequently reduce to the neutral score 0.5. These results delimit the domain of applicability rather than invalidate the approach: TCS is a specialized metric for temporal inconsistencies in political narratives, complementary to, not a replacement for, general-purpose detectors.

%\ag{de scos mai bine in evidenta explicabilitate (inclusiv figura 5.9 di teza)}
% addressed via Fig.~\ref{fig:xai} in Section III-C (new explanation-generation flow figure)
\subsection{Failure Analysis}
The single residual false positive is an accurate article on the Clinton impeachment: the factual-contradiction checker (V4) matches the objects associated with two distinct entities that were legitimately co-involved in the same events, lacking the narrative context to distinguish co-involvement from contradiction. This identifies a structural limitation of object-matching contradiction detection and motivates context-sensitive verification as future work. A negative result is also reported for extraction: a generative relation-extraction pipeline based on REBEL~\cite{rebel2021} was evaluated on the same benchmark and rejected at F1 $=0.038$, mirroring the tested-and-rejected Wikipedia verification level (L4).

The main limitations are the low recall, the synthetic nature of the primary benchmark, the dependence on extraction quality and reference coverage, and the restriction to English-language text, all documented as directions for improvement; an extended 150-article benchmark has been constructed for future validation.

\section{Conclusions}
This paper introduced the Temporal Coherence Score, an interpretable article-level metric for the detection of temporal inconsistencies in the news. 
The metric is computed by a pipeline that combines:  (i) temporal knowledge graphs, (ii) eight internal consistency checkers grounded in interval algebra, and (iii) a five-level hierarchical external verification. At the selected operating point, the system achieves a precision of 0.909 with a single false positive, a profile suited to human-in-the-loop fact-checking assistance, and every verdict is accompanied by a verifiable justification. Evaluation on external datasets explicitly delimits the domain of applicability to temporally grounded political narratives.

Future work targets recall: extending the reference knowledge base beyond US, UK, and EU coverage, refining the extraction of implicitly expressed temporal facts, and introducing context-sensitive contradiction verification to eliminate the residual false positive. Integrating causal temporal reasoning models~\cite{wang2025explainablefakenews,local2025factchecking} is another direction for inconsistencies that exceed interval-based verification. The results indicate that temporal coherence is a viable and explainable signal for a specific class of disinformation, complementing rather than replacing lexical detectors.

The code and the online demo are available after blind review.

\section*{Ethical Considerations}
The benchmark's fake articles contain temporal errors deliberately injected into otherwise factual accounts of real public figures, for the sole purpose of controlled evaluation; this text is not intended for, nor suitable for, circulation outside the research setting. All reference facts derive from Wikidata~\cite{vrandecic2014wikidata}, a public, collaboratively maintained knowledge base, used under its open licensing terms; no personal or private data is involved. Given its precision-oriented operating point, its low recall, and its demonstrated inability to generalize beyond temporally grounded political narratives (Section~\ref{subsec:domain}), the system is intended to assist, not replace, human fact-checkers, and should not be deployed as an autonomous arbiter of an article's veracity, particularly given the reputational risk to public figures of an erroneous flag.

\section*{Acknowledgment}
This work was supported by a grant of the Ministry of Research, Innovation and Digitization, CCCDI-UEFISCDI, project number
PN-IV-P6-6.3-SOL-2024-2-0312 within PNCDI IV.

\balance

\bibliographystyle{IEEEtran}
\bibliography{articol-iccp}

\end{document}